\documentclass[10pt,twocolumn,letterpaper]{article}
\usepackage[pagenumbers]{iccv}
\PassOptionsToPackage{table}{xcolor}

\usepackage[accsupp]{axessibility}
\usepackage{arydshln}
\usepackage{bm}
\usepackage{boldline}
\usepackage{enumitem}
\usepackage{float}
\usepackage{multirow}
\usepackage{nicefrac}
\usepackage{pifont}
\usepackage[most]{tcolorbox}
\usepackage{mathtools}
\usepackage{makecell}
\usepackage{colortbl}               
\usepackage[table]{xcolor}

\definecolor{Urlcolor}{RGB}{251,111,146}
\definecolor{Linkcolor}{RGB}{193,18,31}
\definecolor{CiteColor}{RGB}{32,126,190}
\definecolor{darkgreen}{RGB}{0.0, 0.5, 0.0}
\usepackage[pagebackref,breaklinks,colorlinks,urlcolor=Urlcolor,linkcolor=Linkcolor,citecolor=CiteColor]{hyperref}

\newcommand{\cmark}{\ding{51}}
\newcommand{\xmark}{\ding{55}}

\makeatletter
\renewcommand{\paragraph}{%
    \@startsection{paragraph}{4}%
    {\z@}{-0.5em}{-0.5em}%
    {\normalfont\normalsize\bfseries}%
}
\makeatother

\newcommand{\methodold}{DragAPart\xspace}
\newcommand{\method}{Puppet-Master\xspace}
\newcommand{\dataset}{Objaverse-Animation\xspace}
\newcommand{\datasetF}{Objaverse-Animation-HQ\xspace}

\newcommand{\tablestyle}[2]{\setlength{\tabcolsep}{#1}\renewcommand{\arraystretch}{#2}\centering\footnotesize}

\newcommand\rurl[1]{%
  \href{https://#1}{\nolinkurl{#1}}%
}

\def\paperID{3}
\def\confName{ICCV}
\def\confYear{2025}

\title{Puppet-Master: Scaling Interactive Video Generation as a Motion Prior for Part-Level Dynamics}

\author{Ruining Li
\quad
Chuanxia Zheng
\quad
Christian Rupprecht
\quad
Andrea Vedaldi \\
Visual Geometry Group, University of Oxford \\
{\tt\small \{ruining, cxzheng, chrisr, vedaldi\}@robots.ox.ac.uk}\\[0.1em]
\small\rurl{vgg-puppetmaster.github.io}
}
\togglefalse{iccvpagenumbers}

\begin{document}
\twocolumn[\maketitle\thispagestyle{empty}\vspace{-3em}\begin{center}
\includegraphics[trim=0 0 0 0, clip, width=0.99\linewidth]{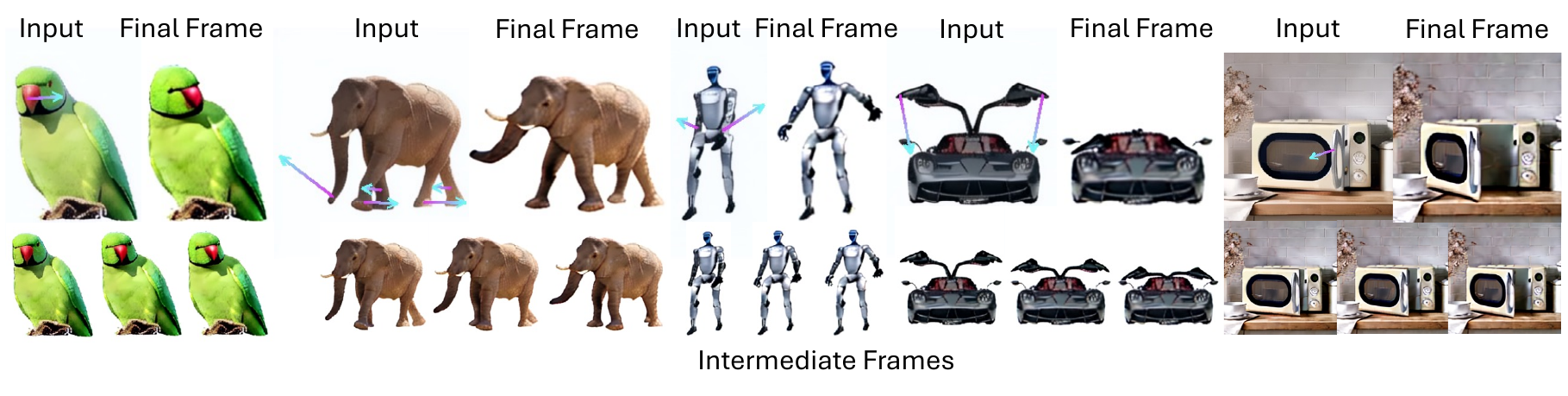}
\end{center}\vspace{-1.5em}
\captionof{figure}{\textbf{\method} generates videos depicting \emph{internal, part-level} motion, prompted by 
one or more drags (arrows).
Fine-tuned solely on our curated synthetic \datasetF dataset,
it generalizes well to real-world scenarios and diverse object categories.
}%
\label{fig:teaser}\bigbreak]
\begin{center}
\large\bfseries Abstract
\end{center}
\textit{We introduce \method, an interactive video generator that captures the \emph{internal, part-level} motion of objects, serving as a proxy for modeling object dynamics universally.
Given an image of an object and a set of ``drags'' specifying the trajectory of a few points on the object, the model synthesizes a video where the object’s parts move accordingly.
To build \method, we extend a pre-trained image-to-video generator to encode the input drags.
We also propose \emph{all-to-first} attention, an alternative to conventional spatial attention that mitigates artifacts caused by fine-tuning a video generator on out-of-domain data.
The model is fine-tuned on \datasetF, a new dataset of curated \emph{part-level} motion clips obtained by rendering synthetic 3D animations.  
Unlike real videos, these synthetic clips avoid confounding part-level motion with overall object and camera motion.  
We extensively filter sub-optimal animations and augment the synthetic renderings with meaningful drags that emphasize the internal dynamics of objects.  
We demonstrate that \method learns to generate part-level motions, unlike other motion-conditioned video generators that primarily move the object as a whole.  
Moreover, \method generalizes well to out-of-domain real images, outperforming existing methods on real-world benchmarks in a \emph{zero-shot} manner.
}

\section{Introduction}%
\label{sec:intro}

In this paper, we introduce \method, an \emph{interactive} video generator that predicts how objects move in response to external stimuli.  
This generator takes as input a single image of an object and a set of sparse \emph{drags}, which specify the motion of selected points on the object.
It then outputs a video of the \emph{part-level} object motion consistent with the drags and the object's internal dynamics (\cref{fig:teaser}).  

We are motivated by the need for AI systems to understand how objects can move and deform in general.  
Researchers have developed countless models of dynamic objects, but most are specific to particular object \emph{types}, such as faces, hands, humans, or quadrupeds~\citep{blanz99a-morphable,romero22embodied,loper15smpl,zuffi173d-menagerie}.  
The few more general models~\citep{tang2022neural} do not make strong assumptions about object types but are difficult to train due to the lack of suitable data (\eg, aligned 3D meshes for \citep{tang2022neural}).  
None of these are good candidates for learning a ``foundation'' model of part-level object dynamics.  
Such a model should be able to express different types of natural object dynamics (\cref{fig:teaser}), such as part articulation, sliding of parts, and soft deformations.  

Recently, video generators trained on millions of videos have been proposed as proxies for ``world models''~\citep{brooks24video, parkerholder2024genie2, agarwal2025cosmos}.  
Any general world model should possess an understanding of object dynamics.  
However, these models, trained on Internet-scale data, still struggle to capture the nuances of \emph{internal, part-level} dynamics.  
Inspired by \methodold~\citep{li2024dragapart}, we consider learning a \emph{conditional} video generator that predicts the \emph{part-level} motion of objects in pixel space in response to sparse motion trajectories.  
This generator takes as input a single image of an object and a set of \emph{drags}, which specify the motion of selected points on the object.
It then outputs a video of the \emph{part-level} object motion consistent with the drags (\cref{fig:teaser}).  

Several authors have already explored incorporating drag-like motion prompts in image or video generation~\citep{blattmann2021ipoke, chen2023motion, pan2023drag, yin2023dragnuwa, li2024generative, wang2023motionctrl, shi2024dragdiffusion, mou2023dragondiffusion, geng2024motion, ling2023freedrag, wu2024draganything, mou2024revideo, li2024imageconductor, geng2024motionprompting, li2024dragapart}.  
Many such works utilize techniques like ControlNet~\citep{zhang23adding} to inject motion control into a pre-trained generator.  
However, when fine-tuned on real-world videos with motion conditions extracted using off-the-shelf trackers, these models often respond to drags by merely shifting or scaling entire objects, failing to capture their internal dynamics, such as a microwave door rotating shut or a fish oscillating its tail (\cref{fig:teaser} and \cref{fig:real}).  
This limitation can be attributed, in part, to the various confounding components inherent in natural videos, including occlusions, background variations, and camera movements, which complicate motion learning and synthesis.  
Hence, the challenge is to encourage video generators to synthesize \emph{internal, part-level} dynamics.  

In this work, we aim to develop a model capable of generating part-level motion, leveraging \emph{synthetic} data that eliminates the confounding factors present in real-world videos and emphasizes part-level dynamics.  
We start from a large-scale pre-trained generator, Stable Video Diffusion (SVD)~\citep{blattmann2023stable}, and show how to repurpose it for motion prediction.  
We make the following \textbf{contributions}.  

First, we introduce new modules into the video generator for effective motion control and improved appearance generation.  
In particular, we incorporate \emph{drag tokens} into cross-attention modules for enhanced conditioning.  
These tokens, regressed from the start and end points of each drag using an encoding function, supplement the \emph{single} image token used in the original SVD, improving spatial awareness in cross-attention.  
In addition, we introduce \emph{all-to-first} attention, which addresses the degradation in appearance quality that often arises when fine-tuning diffusion generators on out-of-distribution datasets~\cite{li2024instant3d, zuo2024videomv, li2024vividzoo}.  
In our design, \emph{all} frames attend to the first one via a variant of self-attention.  
This creates a shortcut that directly propagates information from the clean conditioning frame to the others, preventing the model from getting stuck in local optima.  

Our second contribution is to provide two datasets to learn part-level object motion.  
Both datasets comprise subsets of the 40k animated assets in Objaverse~\citep{deitke22objaverse}.  
Objaverse animations vary in quality:  
while some display realistic object dynamics, others feature objects that  
(i) are static,  
(ii) exhibit simple translations, rotations, or scaling, or  
(iii) move in a physically implausible way.  
We introduce a systematic approach for large-scale animation curation.  
The resulting datasets, \dataset (16k animations) and \datasetF (10k animations), contain progressively higher-quality 3D animations.  
Empirical results show that \datasetF, despite its more modest size, yields a superior model compared to \dataset, highlighting the effectiveness of our data curation strategy.  

With the new curated datasets, we train \emph{\method}, our new video generative model that, given a single image of an object and corresponding drags, generates an animation of the object.  
These animations are faithful to both the input image and the sparse motion trajectories, while exhibiting physically plausible motions at the level of individual object parts (\cref{fig:teaser}).  
Our model works across a diverse set of object categories.  
Empirically, it outperforms prior works on multiple benchmarks.  
We also present ablations to validate our design choices.  
Notably, although our \method is fine-tuned using only synthetic data, it generalizes well to real data without further tuning.
\section{Related Work}%
\label{sec:rel}

\paragraph{Generative models.}

Recent advances in generative models, largely powered by diffusion models~\citep{ho20ddpm, song2019generative, song2021scorebased}, have enabled photorealistic synthesis of images~\citep{ramesh21zero-shot, rombach2022stablediffusion, saharia2022photorealistic} and videos~\citep{ho2022imagen-video, blattmann2023videoldm, girdhar2023emu, blattmann2023stable}, and have been extended to various other modalities~\citep{tevet2022human, lei2023nap}.  
Generation is primarily controlled by a text or image prompt.  
Recent works have explored leveraging these models' prior knowledge through either score distillation sampling~\citep{poole2022dreamfusion, lin2023magic3d, melas2023realfusion, jakab2024farm3d} or fine-tuning on specialized data for downstream applications, such as multi-view images for 3D asset generation~\citep{liu23zero-1-to-3, li2024instant3d, melas-kyriazi2024IM3d, zheng2023free3D, voleti2024sv3d, gao2024cat3d}.  

\begin{figure*}[tb!]
\centering
\includegraphics[width=0.8\linewidth]{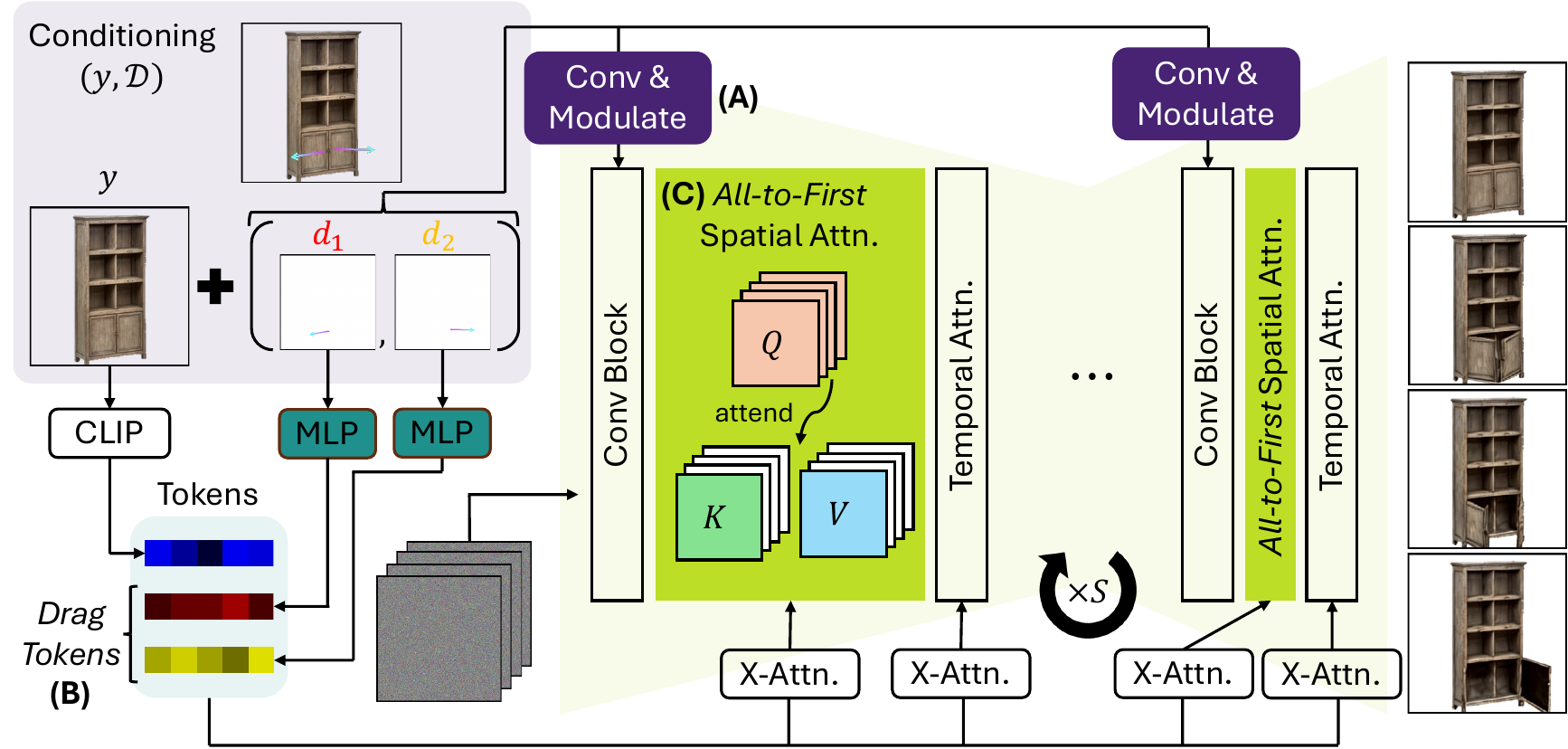}
\caption{\textbf{Architectural Overview of \method}.
To enable precise drag conditioning, we first modify the original latent video diffusion architecture (\cref{sec:method_svd}) by
(\textbf{A}) adding adaptive layer normalization modules to modulate the internal diffusion features and
(\textbf{B}) adding cross attention with \emph{drag tokens} (\cref{sec:method_drag}).
Furthermore, to ensure high-quality appearance and background,
we introduce (\textbf{C}) \emph{all-to-first} attention, a drop-in replacement for the spatial self-attention modules, where every video frame attends the first one (\cref{sec:method_attn}).
}
\label{fig:framework}
\end{figure*}

\paragraph{Video generation for motion.}

Modeling object motion often relies on pre-defined shape models, \eg, SMPL~\citep{loper15smpl} for humans and SMAL~\citep{zuffi173d-menagerie} for quadrupeds, which are limited to specific categories.  
Videos can capture general object dynamics~\citep{yang2024video, brooks24video}, but existing video generators trained on Internet-scale videos often produce incoherent motion.  
Researchers have explored controlling video generation with motion trajectories.  
{}\cite{teng2023drag} extends the framework of~\cite{pan2023drag} to videos, relying on the motion prior of pre-trained video generators, which may not produce high-quality results.  
Training-based methods \emph{learn} drag-based control using ad-hoc training data.  
Early efforts~\citep{blattmann2021ipoke, davtyan2023learn} train variational autoencoders or diffusion models to synthesize videos with objects in motion, conditioned on sparse motion trajectories derived from optical flow.  
{}\cite{li2024generative} uses a Fourier-based motion representation for natural, oscillatory dynamics like trees and candles, generating motion with a diffusion model.  
DragNUWA~\citep{yin2023dragnuwa} and others~\citep{wang2023motionctrl, wu2024draganything, mou2024revideo, li2024imageconductor, geng2024motionprompting} fine-tune pre-trained video generators on large video datasets augmented with motion prompts obtained from off-the-shelf trackers, enabling drag-based control in open-domain video generation.  
However, these methods do \emph{not} control motion at the object part level, as their training data entangles multiple factors, making it challenging to model part-level motion.  
Other works leverage the motion prior of video generative models for 4D generation tasks~\citep{liang2024diffusion4d, zhang20244diffusion, jiang2024animate3d, xie2024sv4d}, but they lack dragging control.
\section{Method}%
\label{sec:method}

Given the initial state of an object, represented by an image $y$, and one or more drags
$
\mathcal{D}=\left \{ d_k \right \}_{k=1}^K
$,
our goal is to synthesize a video $\mathcal{X}=\{x_i\}_{i=1}^{N}$ sampled from the distribution
$
\mathcal{X}
\thicksim
\mathbb{P}(x_1, x_2, \dots, x_N|y,\mathcal{D})
$,
where $N$ is the number of video frames.
The distribution $\mathbb{P}$ should generate a physically plausible \emph{part-level} animation of the object that responds to the drags.
For generalizability, we leverage a ``foundation'' video generator, Stable Video Diffusion (SVD,~\cref{sec:method_svd})~\citep{blattmann2023stable}, which has a general understanding of motion, acquired by training on millions of Internet videos.

In this section, we describe how to fine-tune such a pre-trained video generator to enable part-level motion control of objects.
There are two main challenges.
First, the drag conditioning must be injected into the video generation pipeline to facilitate efficient learning and accurate, time-consistent motion control.
This must be done without strongly interfering with the internal pre-trained video representation.
Second, na{\"\i}vely fine-tuning a pre-trained video diffusion model can result in artifacts such as cluttered backgrounds~\citep{li2024instant3d}, particularly when the fine-tuning data distribution differs significantly from that of the pre-training data.
To address these challenges, in~\cref{sec:method_drag}, we first introduce a novel mechanism to inject the drag condition $\mathcal{D}$ into the video diffusion model.
Then, in \cref{sec:method_attn}, we improve the quality of the generated videos by introducing an \emph{all-to-first} attention mechanism, which reduces artifacts like background clutter.
While we build on SVD, these techniques should be easily portable to other video generators based on diffusion.

\subsection{Preliminaries: Stable Video Diffusion}%
\label{sec:method_svd}

SVD is an image-conditioned video generator based on diffusion, implementing a denoising process in latent space.
It utilizes a variational autoencoder (VAE) $(E, D)$, where the encoder $E$ maps the video frames to the latent space, and the decoder $D$ reconstructs the video from the latent codes.
During training, given a pair $(\mathcal{X}, y)$ formed by a video $\mathcal{X}=x^{1:N}$ and the corresponding image prompt $y$, one first obtains the latent code as
$
z_0^{1:N}
=
E(x^{1:N})
$,
and then adds Gaussian noise
$
\epsilon
\thicksim
\mathcal{N}(0,\bm{I})
$,
obtaining the progressively more noised codes
\begin{equation}
\label{eq:markov}
z_t^{1:N}
=
\sqrt{\bar{\alpha}_t}z_0^{1:N} +
\sqrt{1-\bar{\alpha}_t}\epsilon^{1:N},
~~~
t=1,\dots,T.
\end{equation}
This uses a pre-defined noising schedule $\bar{\alpha}_0=1, \dots, \bar{\alpha}_T=0$.
The denoising network $\epsilon_\theta$ is trained to reverse this noising process by optimizing the objective function:
\begin{equation}\label{eq:objective}
    \min_\theta
    \mathbb{E}_{(x^{1:N},y), t, \epsilon^{1:N}\sim\mathcal{N}(0,\bm{I})}
    \left[
    \| \epsilon^{1:N} - \epsilon_\theta(z_t^{1:N}, t, y) \|^2_2
    \right].
\end{equation}
Here, $\epsilon_\theta$ uses the same U-Net architecture as~\cite{blattmann2023videoldm}, inserting temporal convolution and temporal attention modules after the spatial modules used by~\cite{rombach2022stablediffusion}.
The image conditioning is achieved via
(1) cross-attention with the CLIP~\cite{radford2021learning} embedding of the reference frame $y$; and
(2) concatenating the encoded reference image $E(y)$ channel-wise to $z_t^{1:N}$ as the input of the network $\epsilon_\theta$.
After $\epsilon_\theta$ is trained, the model generates a video $\hat{\mathcal{X}}$ prompted by $y$ via iterative denoising from pure Gaussian noise $z_T^{1:N}\sim\mathcal{N}(0,\bm{I})$, followed by VAE decoding:
$
\hat{\mathcal{X}}
=
\hat{x}^{1:N}
=
D(z_0^{1:N})
$.

\subsection{Adding Drag Control to Video Diffusion Models}%
\label{sec:method_drag}

Here, we show how to add the drags $\mathcal{D}$ as an additional input to the denoiser $\epsilon_\theta$ for part-level motion control.
This is achieved by introducing an encoding function for the drags $\mathcal{D}$ and by extending the SVD architecture to inject the resulting code into the network.
The model is then fine-tuned using videos combined with corresponding drag prompts in the form of training triplets $(\mathcal{X}, y, \mathcal{D})$.
We summarize the key components of the model below and refer the reader to \cref{sec:supp_enc} for more details.

\paragraph{Drag encoding.}

Let $\Omega$ be the spatial grid
$
\left \{ 1,\ldots,H \right \}
\times
\left \{ 1,\ldots,W \right \}
$,
where $H\times W$ is the resolution of the video.
A \emph{drag} $d_k$ is a tuple $(u_k, v_k^{1:N})$ specifying that the drag starts at location $u_k\in\Omega$ in the reference image $y$ and lands at locations $v_k^n\in\Omega$ in subsequent frames.
To encode a set of drags
$
\mathcal{D} = \left \{ d_k \right \}_{k=1}^K
$,
where $K \leq K_{\max}=5$, we use the multi-resolution encoding of~\cite{li2024dragapart}.
Each drag $d_k$\footnote{With a slight abuse of notation, we assume $d_k\in \Omega^N$, as $u_k=v_k^1$ and hence $v_k^{1:N}\in\Omega^N$ fully describes $d_k$.} is fed to a hand-crafted encoding function
$
\operatorname{enc}(\cdot, s):
\Omega^{N} \mapsto \mathbb{R}^{N\times s\times s\times c}
$,
where $s$ is the desired encoding resolution.
The encoding function captures the state of the drag in each frame.
Specifically, each slice
$
\operatorname{enc}(d_k, s)[n]
$
encodes
(1) the drag's starting location $u_k$ in the reference image,
(2) its intermediate location $v_k^n$ in the $n$-th frame, and
(3) its final location $v_k^{N}$ in the last frame.
The $s \times s$ map $\operatorname{enc}(d_k, s)[n]$ is filled with values $-1$ except at the three locations $u_k$, $v_k^n$, and $v_k^N$, which are encoded using $c=6$ channels.
Finally, we obtain the encoding
$
\mathcal{D}_{\operatorname{enc}}^s\in
\mathbb{R}^{N\times s\times s\times c K_{\max}}
$
of $\mathcal{D}$ by concatenating the encodings of the $K$ individual drags, filling extra channels with $-1$ if $K < K_{\max}$.
The encoding function is further detailed in~\cref{sec:supp_enc}.

\paragraph{Drag modulation.}

The SVD denoiser $\epsilon_\theta$ comprises a sequence of U-Net blocks computing feature maps
$
f_s \in \mathbb{R}^{N\times s\times s\times C}
$
at different resolutions $s$.
We update each feature $f_s$ based on the drag encoding $\mathcal{D}_{\text{enc}}^s$ using an adaptive normalization module~\citep{perez2018FiLM}, \ie,
\begin{equation}
    f_s \leftarrow f_s \otimes 
    (\mathbf{1} + \gamma_s(\mathcal{D}_{\text{enc}}^s)
    ) + \beta_s(\mathcal{D}_{\text{enc}}^s),
\end{equation}
where $\otimes$ denotes element-wise multiplication.
$
\gamma_s
$
and 
$\beta_s \in \mathbb{R}^{N\times s\times s\times C}
$
are the \emph{scale} and \emph{shift} terms regressed from the drag encoding $\mathcal{D}_{\text{enc}}^s$.
We use convolutional layers to embed $\mathcal{D}_{\text{enc}}^s$ from dimension $c K_{\max}$ to the target dimension $C$.
We empirically find that this mechanism provides better conditioning than using only a single shift term with \emph{no} scaling as in DragAPart~\cite{li2024dragapart} (see ablation in \cref{tab:ablations}).

\paragraph{Drag tokens.}

In addition to drag modulation conditioning,
we also condition the network $\epsilon_\theta$ via SVD's built-in cross-attention modules.
These modules attend to a \emph{single} key-value pair obtained from the CLIP~\citep{radford2021learning} encoding of the reference image $y$,
and thus degenerate to a global bias term with \emph{no} spatial awareness~\citep{sobol2024zero}.
In contrast, we concatenate to the CLIP token additional \emph{drag tokens} so that cross-attention is non-trivial.
We use multi-layer perceptrons (MLPs) to regress an additional key-value pair from \emph{each} drag $d_k$.
The MLPs take the origin $u_k$ and terminations $v_k^n$ and $v_k^N$ of $d_k$, along with the internal diffusion features sampled at these locations, which are shown to contain semantic information~\citep{baranchuk2021label}, as inputs.
Overall, the cross-attention modules have $1 + K$ key-value pairs ($1$ is the original image CLIP embedding).

\subsection{All-to-First Attention}%
\label{sec:method_attn}

In our preliminary experiments, we noted that the background of the generated videos does not match the input image $y$ well, often appearing grayer.
Instant3D~\citep{li2024instant3d} reported a similar problem when generating multiple views of a 3D object, which they addressed via careful noise initialization.
\citep{zuo2024videomv} and \citep{li2024vividzoo} directly constructed training videos with a gray background, which might mitigate the issue visually.

To investigate this issue, we prompt the pre-trained SVD with an image of resolution $256\times256$ (the resolution used during fine-tuning).
As shown in~\cref{sec:supp_svd}, SVD, originally trained on $1024\times576$ videos, fails to generalize to very different resolutions.
We hypothesize that the suboptimal results obtained through fine-tuning arise from the significant discrepancy between the distribution of SVD's training videos and that of our fine-tuning videos, both in terms of resolution and visual content.
However, we noticed that the first frame of each generated video is spared from appearance degradation (\cref{fig:ablations}),
as the model effectively replicates the reference image.
This suggests that the initial frame serves as a stable foundation for the subsequent frames.
To leverage this stability, we propose an \emph{all-to-first} attention mechanism,
which introduces a \emph{shortcut} from each noised frame to the first frame via attention.
\begin{figure*}[tb!]
    \centering
    \includegraphics[width=\textwidth]{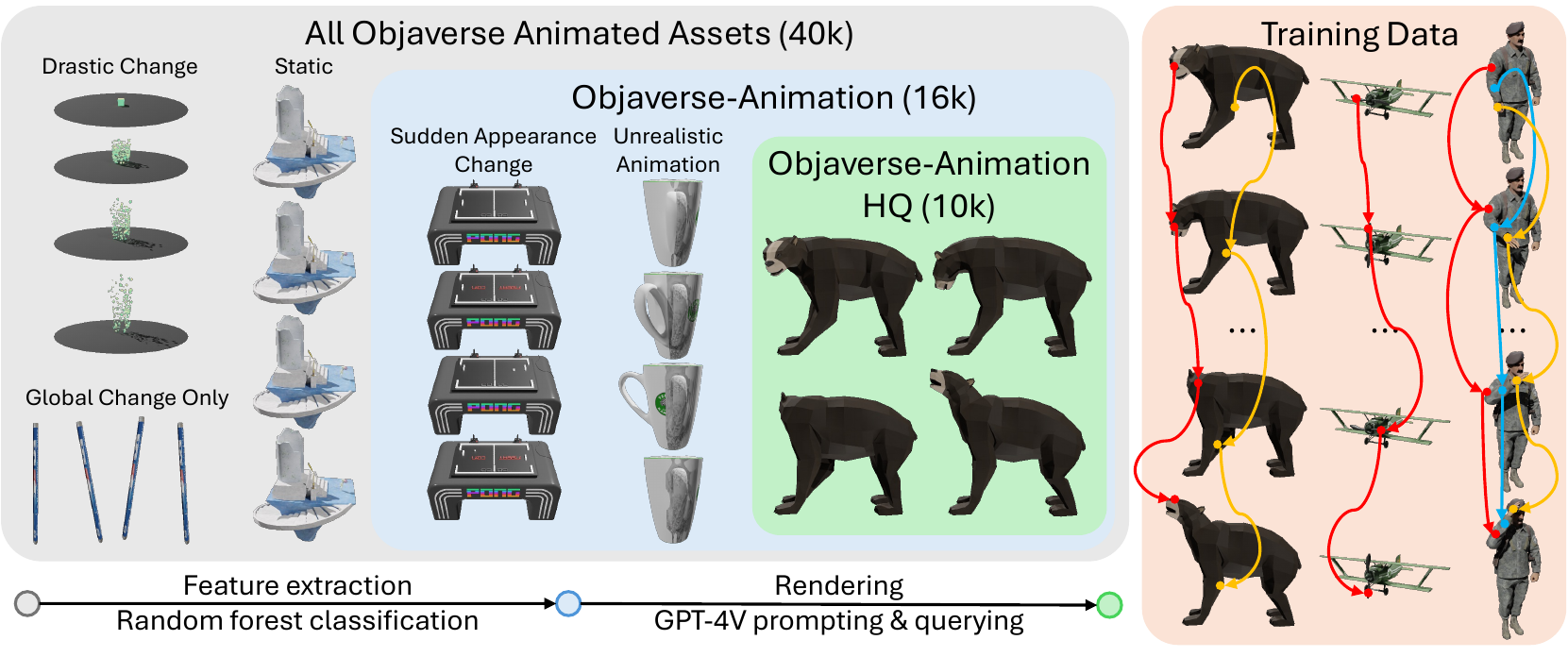}
    \vspace{-20pt}
    \caption{\textbf{Data Curation}.
    We propose two strategies to filter the animated assets in Objaverse, resulting in \dataset ($16$k) and \datasetF ($10$k) of varying levels of curation, from which we construct the training data of \method by sampling sparse motion trajectories and projecting them to 2D as drags.
    }%
    \label{fig:dataset}
\end{figure*}

Previous works~\citep{watson2023novel, cao_2023_masactrl, weng2023consistent123} have shown that attention between the noised branch and the reference branch improves generation quality for image editing and novel view synthesis tasks.
In our \emph{all-to-first} attention, each noised frame attends to the first (reference) frame.
We implement this attention by having each frame query the key and value of the first frame, modifying all self-attention layers in the denoising U-Net $\epsilon_\theta$.
More specifically, denoting the query, key, and value tensors as $Q, K$, and
$
V\in
\mathbb{R}^{N\times s\times s\times C}
$,
we discard the key and value tensors of non-first frames,
and compute the spatial attention $A_i$ of the $i$-th frame as follows:
\begin{equation}
    A_i = \operatorname{softmax}
    \left(
        \frac{ 
          \operatorname{flat}(Q[i])
          \operatorname{flat}(K[0])^\top
        }  
        {\sqrt{D}}
    \right)
    \operatorname{flat}(V[0]),
\end{equation}
where
$
\operatorname{flat}(\cdot) :
\mathbb{R}^{s\times s\times C}
\mapsto
\mathbb{R}^{L\times C}
$
flattens the spatial dimensions to get $L=s \times s$ tokens for attention.
The benefit is two-fold:
first, this shortcut to the first frame allows subsequent frames to access non-degraded appearance details of the reference image directly.
Second, combined with the proposed drag encoding (\cref{sec:method_drag}), which specifies the origin $u_k$ at the first frame for \emph{every} frame, all-to-first attention enables the latent pixel corresponding to the drag termination (\ie, $v_k^n$) to more easily attend to the latent pixel corresponding to the drag origin in the first frame, thereby facilitating learning.

\section{Curating Data for Part-Level Object Motion}%
\label{sec:dataset}

For training, we require a video dataset that captures the motion of objects at the level of parts.
Motion-conditioned video generators fine-tuned using real-world video datasets often confuse part-level motion with object-level motion.
It is challenging to curate a high-quality video dataset from Internet videos that features exclusively part-level dynamics.
The work of~\cite{li2024dragapart} instead used renderings of synthetic 3D objects and their corresponding part annotations obtained from GAPartNet~\citep{geng2023gapartnet}.
Unfortunately, this dataset requires manual annotation and animation of 3D object parts, which limits its scale.
We instead turn to Objaverse~\citep{deitke22objaverse}, a large-scale 3D dataset of $800$k models created by 3D artists, among which $40$k are animated.
In this section, we introduce a pipeline to extract suitable training videos from these animated assets, together with corresponding drags $\mathcal{D}$.

\begin{table*}[tb!]
    \newcommand{\STAB}[1]{\begin{tabular}{@{}c@{}}#1\end{tabular}}
    \tablestyle{2.0pt}{1.0}
    \centering
    \begin{tabular}{@{}l ccc ccccc c cccc@{}}
        \toprule
        \multirow{2}{*}{\textbf{Method}} & \multirow{2}{*}{Video} & \multirow{2}{*}{\shortstack{\!Base \\ Model}} & \multirow{2}{*}{\shortstack{\!Training \\ Data}} & \multicolumn{5}{c}{\textbf{Drag-a-Move}} & & \multicolumn{4}{c}{\textbf{Human3.6M}} \\
        \cline{5-9} \cline{11-14}
        & & & &  PSNR$\uparrow$   & SSIM$\uparrow$   & LPIPS$\downarrow$   & FVD$\downarrow$   & \textbf{M}otion \textbf{E}rror$\downarrow$ & & PSNR$\uparrow$   & SSIM$\uparrow$   & LPIPS$\downarrow$   & FVD$\downarrow$ \\
        \midrule
        DragNUWA~\cite{yin2023dragnuwa}                     & \cmark & SVD~\cite{blattmann2023stable} & \makecell{WebVid + \\ \emph{Internal}} & $20.09$            & $0.874$            & $0.172$            & $281.49$            & $17.55 / 15.41$        & & \cellcolor[HTML]{FCE5CD}$17.52$ & \cellcolor[HTML]{F4CCCC}$0.878$   & \cellcolor[HTML]{FCE5CD}$0.158$  & \cellcolor[HTML]{FCE5CD}$466.91$ \\
        DragAnything~\cite{wu2024draganything}                   & \cmark & SVD~\cite{blattmann2023stable} & VIPSeg & $16.71$            & $0.799$            & $0.296$            & $468.46$            & $16.09 / 23.21$       & & $13.29$            & $0.767$            & $0.305$               & $768.63$             \\
        Image Conductor~\cite{li2024imageconductor}                & \cmark & AnimateDiff~\cite{guo2023animatediff} & \makecell{WebVid + \\ RealEstate10K} & $9.20$             & $0.548$            & $0.585$            & $1138.89$           & $20.09 / 27.51$        & & $8.02$             & $0.467$            & $0.628$               & $1957.33$            \\
        \midrule
        DragAPart~\cite{li2024dragapart} & & & & & & & & & & & & & \\
        \textcolor{gray}{--- \emph{Original}} & \textcolor{gray}{\xmark} & \textcolor{gray}{SD}~\cite{rombach2022stablediffusion} & \textcolor{gray}{Drag-a-Move} & \textcolor{gray}{$23.41$} & \textcolor{gray}{$0.925$} & \textcolor{gray}{$0.085$} & \textcolor{gray}{$180.27$} & \textcolor{gray}{$14.17 / 3.71$} & & \textcolor{gray}{$15.14$} & \textcolor{gray}{$0.852$} & \textcolor{gray}{$0.197$} & \textcolor{gray}{$683.40$} \\
        --- \emph{Re-Trained}           & \xmark & SD~\cite{rombach2022stablediffusion} & Ours & \cellcolor[HTML]{FCE5CD}$23.78$ & \cellcolor[HTML]{F4CCCC}$0.927$  & \cellcolor[HTML]{F4CCCC}$0.082$  & \cellcolor[HTML]{F4CCCC}$189.10$  & \cellcolor[HTML]{FCE5CD}$14.34 / 3.73$ & & $15.25$            & $0.860$            & $0.188$               & $549.64$             \\
        \midrule
        \textbf{\method (ours)}                        & \cmark & SVD~\cite{blattmann2023stable} & Ours & \cellcolor[HTML]{F4CCCC}$24.41$   & \cellcolor[HTML]{F4CCCC}$0.927$   & \cellcolor[HTML]{FCE5CD}$0.085$ & \cellcolor[HTML]{FCE5CD}$246.99$ & \cellcolor[HTML]{F4CCCC} $12.21 / 3.53$ & & \cellcolor[HTML]{F4CCCC}$17.59$   & \cellcolor[HTML]{FCE5CD}$0.872$ & \cellcolor[HTML]{F4CCCC}$0.155$      & \cellcolor[HTML]{F4CCCC}$454.76$   \\
        \bottomrule
    \end{tabular}
    \caption{\textbf{Comparisons} with DragNUWA~\cite{yin2023dragnuwa}, DragAnything~\cite{wu2024draganything}, Image Conductor~\cite{li2024imageconductor} and DragAPart~\cite{li2024dragapart} on the Drag-a-Move and Human3.6M datasets.
    Our model has \emph{not} been trained on Human3.6M or any other real video dataset.
    Colors denote best and second best.}%
    \label{tab:sota}
\end{table*}

\paragraph{Identifying animations.}

While Objaverse~\citep{deitke22objaverse} has $40$k assets labeled as animated, not all animations are useful for our purposes (\cref{fig:dataset}).
Notably, in some, the objects remain static throughout the sequence, while others feature drastic changes in the objects' positions or even their appearances.
Therefore, our initial step is to filter out unsuitable animations.
To do so, we extract a sequence of aligned point clouds from each animated model and calculate several metrics for each sequence, including:
(1) the dimensions and location of the bounding box encompassing the entire motion clip,
(2) the size of the largest bounding box for the point cloud at any single timestamp, and
(3) the mean and maximal displacement of all points throughout the sequence.
Using these metrics, we fit a random forest classifier---trained on a subset of Objaverse animations with manually labeled decisions---to determine whether an animation should be included in the training set.
This filtering excludes many assets that exhibit imperceptibly little or overly dramatic motions and results in a subset of $16$k animations, which we dub \dataset.

Further investigation reveals that this subset still contains assets with highly artificial motion, which do not mimic real-world dynamics (\cref{fig:dataset}).
To avoid such unrealistic dynamics leaking into our synthesized videos, we leverage the multi-modal understanding capability of GPT-4V~\citep{openai23gpt4} to assess motion realism.
Specifically, for each animated 3D asset in \dataset, we fix the camera at the front view and render four images at timestamps corresponding to the four quarters of the animation.
We prompt GPT-4V to determine if the motion depicted is sufficiently realistic to qualify for use in training.
This filtering mechanism excludes another $6$k animations, yielding a subset of $10$k animations, which we dub \datasetF.

\paragraph{Sampling drags.}

The goal of drag sampling is to produce a sparse set of drags
$
\mathcal{D} = \left \{ d_k \right \}_{k=1}^K
$, 
where each drag $d_k\coloneqq (u_k, v_k^{1:N})$ tracks a point $u_k$ on the asset in pixel coordinates throughout the $N$ frames of rendered videos.
To encourage the video generator to learn a meaningful motion prior, the set should ideally be both \emph{minimal} and \emph{sufficient}:
each group of independently moving parts should have \emph{one} and \emph{only one} drag corresponding to its motion trajectory, similar to Drag-a-Move~\citep{li2024dragapart}.
For instance, there should be separate drags for different drawers of the same piece of furniture, as their motions are independent, but not for a drawer and its handle, as in this case, the motion of one \emph{implies} that of the other.
However, Objaverse~\citep{deitke22objaverse} lacks the part-level annotation to enforce this property.
To partially overcome this, we find that some Objaverse assets are constructed in a bottom-up manner, consisting of multiple sub-models that align well with semantic parts.
For these assets, we sample one drag per sub-model;
for the rest, we sample a random number of drags in total.
For each drag, we first sample a 3D point on the visible part of the model (or sub-model) with probability proportional to the point's total displacement across $N$ frames, and then project its ground-truth motion trajectory $p_1, \dots, p_N\in \mathbb{R}^3$ to pixel space to obtain $d_k$.
Once all $K$ drags are sampled, we apply a post-processing procedure to ensure that each pair of drags is sufficiently distinct, \ie, for $i\neq j$, we randomly remove one of $d_i$ and $d_j$ if $\|v_i^{1:N} - v_j^{1:N}\|^2_2\leq \delta$, where $\delta$ is a threshold we empirically set to $20N$ for $256\times 256$ renderings.

\section{Experiments}%
\label{sec:exp}

The main goal of our experiments is to show that fine-tuning pre-trained video diffusion models on a high-quality \emph{synthetic} dataset, curated to emphasize \emph{part-level} motion, enables them to generate realistic internal dynamics of \emph{real-world} objects, outperforming counterpart models fine-tuned on real videos.
To this end, we demonstrate qualitative and quantitative improvements over prior works and excellent generalization to real cases in \cref{sec:exp_main}.
The design choices that led to \method are ablated and discussed in \cref{sec:exp_ablations}.
In \cref{sec:exp_data}, we show the effectiveness of the data curation strategy from \cref{sec:dataset}.
Please refer to \cref{sec:supp_exp} for implementation details.

\subsection{Experiment Settings}%
\label{sec:exp_setting}
  
\paragraph{Datasets.}
\method is trained on a combined synthetic dataset of Drag-a-Move~\citep{li2024dragapart} and \datasetF (\cref{sec:dataset}).
For evaluation, we assess its effectiveness using the test set of Drag-a-Move and real data from
Human3.6M~\citep{h36m_pami},
Amazon-Berkeley Objects~\citep{collins2022abo},
and CC-licensed web images in a \emph{zero-shot} manner (\ie, without tuning on real data).
For quantitative evaluation, our test set contains $100$ videos each from Drag-a-Move and Human3.6M, following~\cite{li2024dragapart}.

\paragraph{Metrics.}
For quantitative results,
we report the standard video quality metrics,
including per-frame PSNR, SSIM, LPIPS~\cite{zhang18the-unreasonable}, and FVD~\cite{unterthiner2019fvd}.
To better evaluate the model's ability to capture \emph{part-level} dynamics, we introduce and report another motion-based metric dubbed \textbf{M}otion \textbf{E}rror, or \textbf{ME} for short, which is computed as the L2 distance between the tracks estimated from the generated and ground-truth videos (using~\citep{karaev2023cotracker}).
In \cref{tab:sota}, we report two \textbf{ME} variants: the first (\emph{before} the slash) is averaged among the origins of drags only, \ie,
$
\left \{ u_k\right \}_{k=1}^K
$,
while the second (\emph{after} the slash) is averaged among all object foreground points.
If the generated videos depict part-level dynamics, the second value should be much \emph{smaller} than the first.
This is because, in such videos, motion is restricted to the parts activated by the drags; other parts that are not required to move remain static, which matches the ground truths and reduces the overall error.

\subsection{Main Results}%
\label{sec:exp_main}

\begin{figure}[tb!]
\centering
\includegraphics[width=\columnwidth]{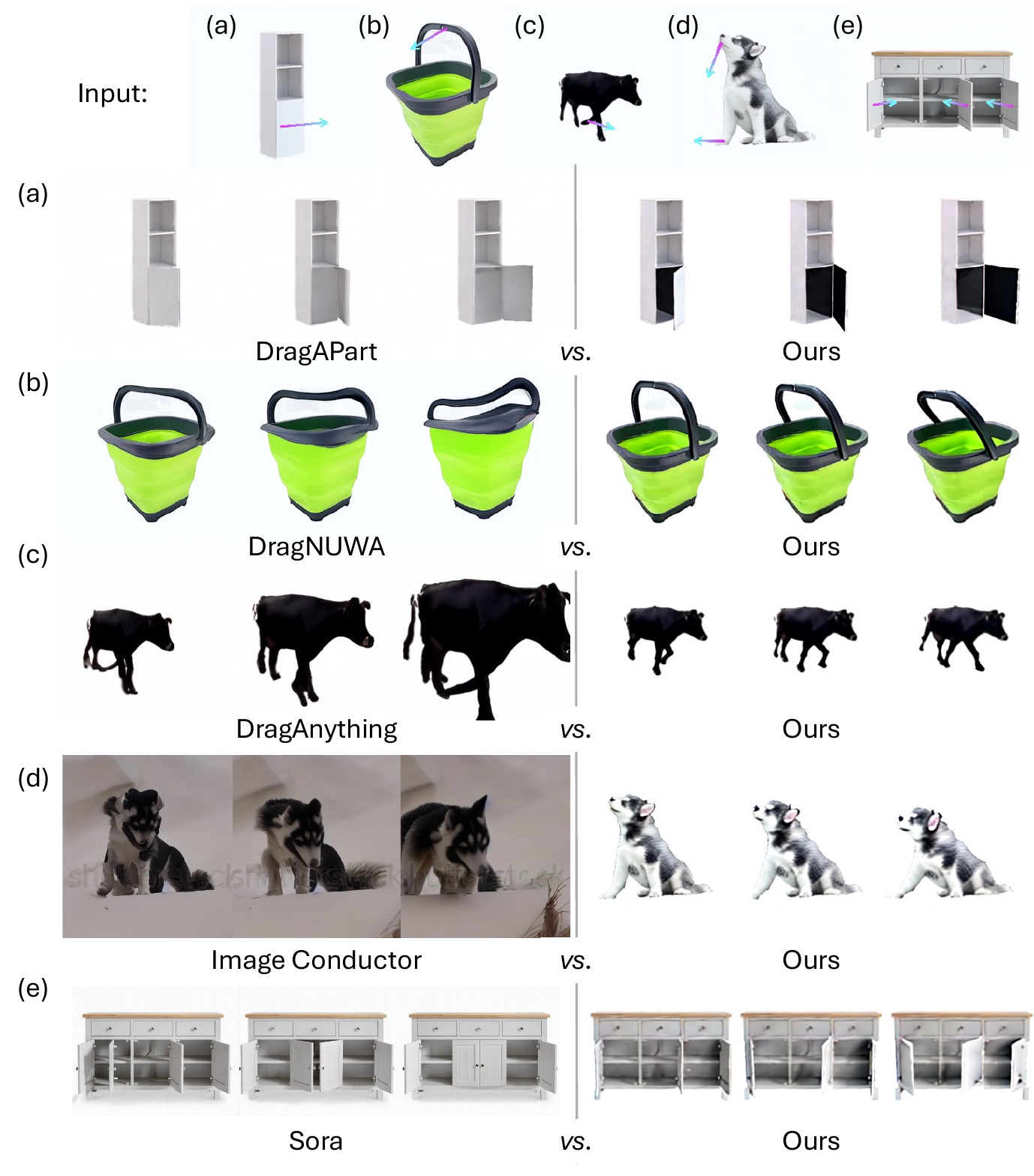}
\vspace{-20pt}
\caption{\textbf{Qualitative Comparison} on real images.
The videos generated by \method are more realistic and capture nuanced part-level dynamics.
}%
\label{fig:comparison}
\end{figure}
\begin{figure}[tb!]
    \centering
    \includegraphics[width=\columnwidth]{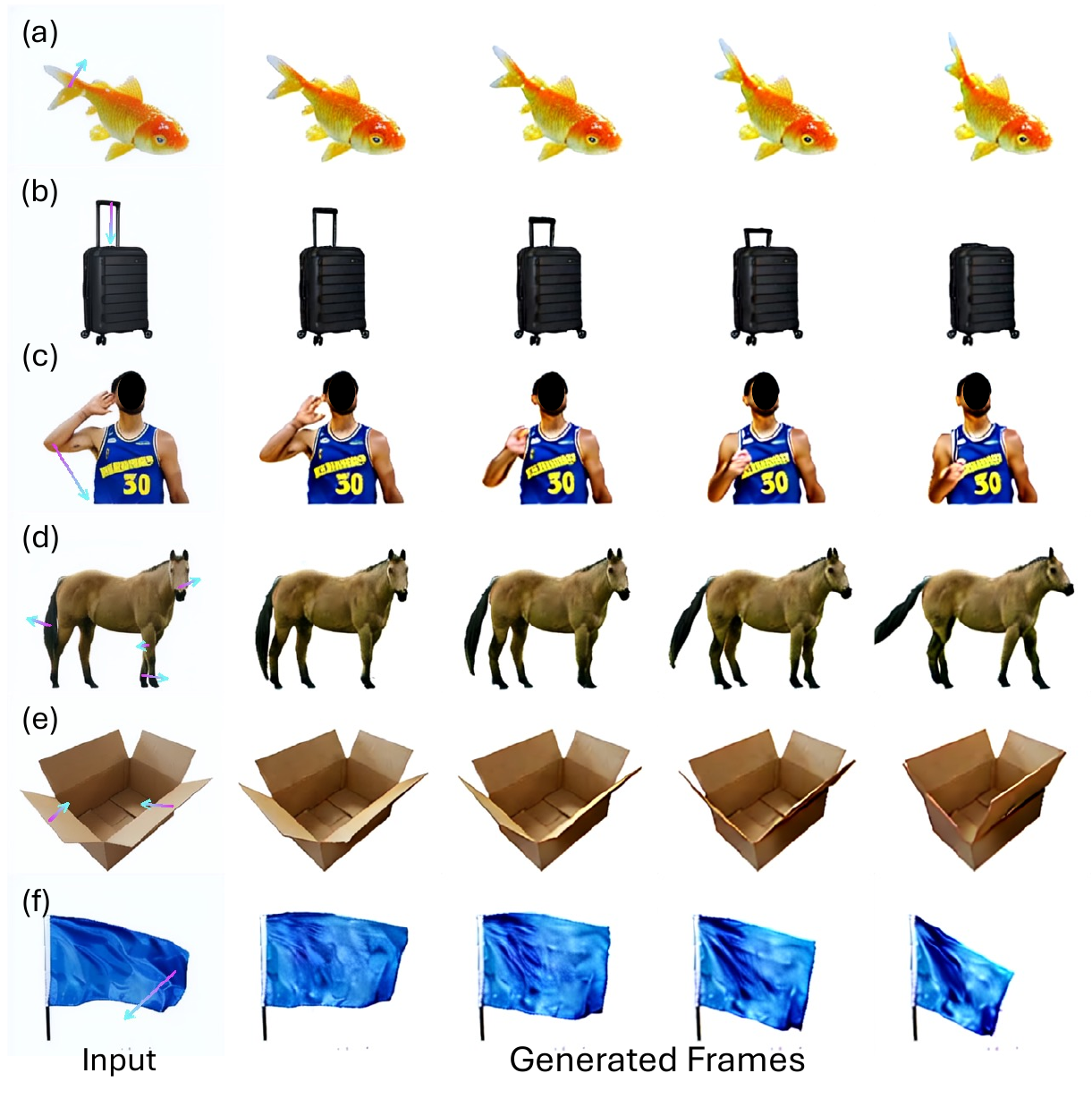}
    \caption{\textbf{Results} on \emph{real} images. The generated videos faithfully adhere to the input drags and exhibit motions representative of the underlying categories, including humans, animals, and both articulated and softly deformable man-made objects.
    In addition, the model learns motion correlations among multiple parts: in (d), without explicitly prompting the rear legs, all four move in sync, while in (e), where the top flaps move independently, only the dragged ones are animated.
    }%
    \label{fig:real}
\end{figure}

\paragraph{Quantitative comparison.}

In \cref{tab:sota}, we compare \method to three state-of-the-art motion-conditioned video generators: DragNUWA~\citep{yin2023dragnuwa}, DragAnything~\citep{wu2024draganything}, and Image Conductor~\cite{li2024imageconductor}, all trained on real data.
On the Drag-a-Move test set, our model consistently outperforms previous models across all metrics.
Interestingly, among the four video generators in \cref{tab:sota},
only \method achieves a markedly better score in the second \textbf{ME} metric.
This highlights \method's superior capability in capturing \emph{part-level} motion dynamics,
while DragNUWA, DragAnything, and Image Conductor predominantly induce whole-object movements, so many points incur large errors.

To assess cross-domain generalizability, we evaluate \method on Human3.6M~\cite{h36m_pami}, an unseen dataset captured in the real world.
On this out-of-domain test set, \method outperforms prior models on most metrics, despite \emph{not} being fine-tuned on any real videos.

We also report the metrics of DragAPart~\citep{li2024dragapart}, a drag-conditioned \emph{image} generator for part-level motion.
The original DragAPart was trained only on the Drag-a-Move dataset.
For fairness, we fine-tune it with the identical data setting as \method, and evaluate the performance of both checkpoints (\textcolor{gray}{\emph{Original}}\footnote{\textcolor{gray}{\emph{Original}} is not ranked as it is trained on single-category data only and hence not an open-domain generator.} and \emph{Re-Trained} in \cref{tab:sota}).
The videos are obtained from $N$ independently generated frames conditioned on gradually extending drags.
While its samples exhibit high visual quality in individual frames, they lack temporal smoothness, characterized by abrupt transitions and discontinuities in movement, resulting in a larger motion error\footnote{FVD is \emph{not} an informative metric for motion quality.
Prior works~\citep{ge2024content, watson20244dim} noted that FVD is biased towards the quality of individual frames and does \emph{not} sufficiently account for motion.}
(\cref{fig:comparison}a).
Furthermore, DragAPart fails to generalize to out-of-domain cases (\cref{tab:sota} Human3.6M), as its base model, Stable Diffusion, was not trained on videos and lacks inherent motion priors.

\paragraph{Qualitative comparison.}

We compare samples generated by \method and prior models in \cref{fig:comparison}.
In addition to the baselines in \cref{tab:sota}, we also compare with Sora~\cite{videoworldsimulators2024}, a commercial video generator with text and keyframe control.
DragAPart, which builds on an image generator, produces samples that lack motion consistency across frames (\cref{fig:comparison}a).
Other video generators \emph{cannot} generate part-level dynamics, introducing unrealistic distortions (\cref{fig:comparison}be) or scaling or moving the entire object (\cref{fig:comparison}cd).
This includes Sora, which has been trained with orders of magnitude more data and compute, suggesting that uncurated Internet videos may not be an efficient source for learning the internal motion of objects.
By contrast, fine-tuned solely on synthetic 3D renderings, \method generates dynamics that are physically plausible, faithful to the input images and drags, and generalizes to real cases.
More examples generated by \method can be found in \cref{fig:real}.

\begin{figure}[tb!]
    \centering
    \includegraphics[width=\columnwidth]{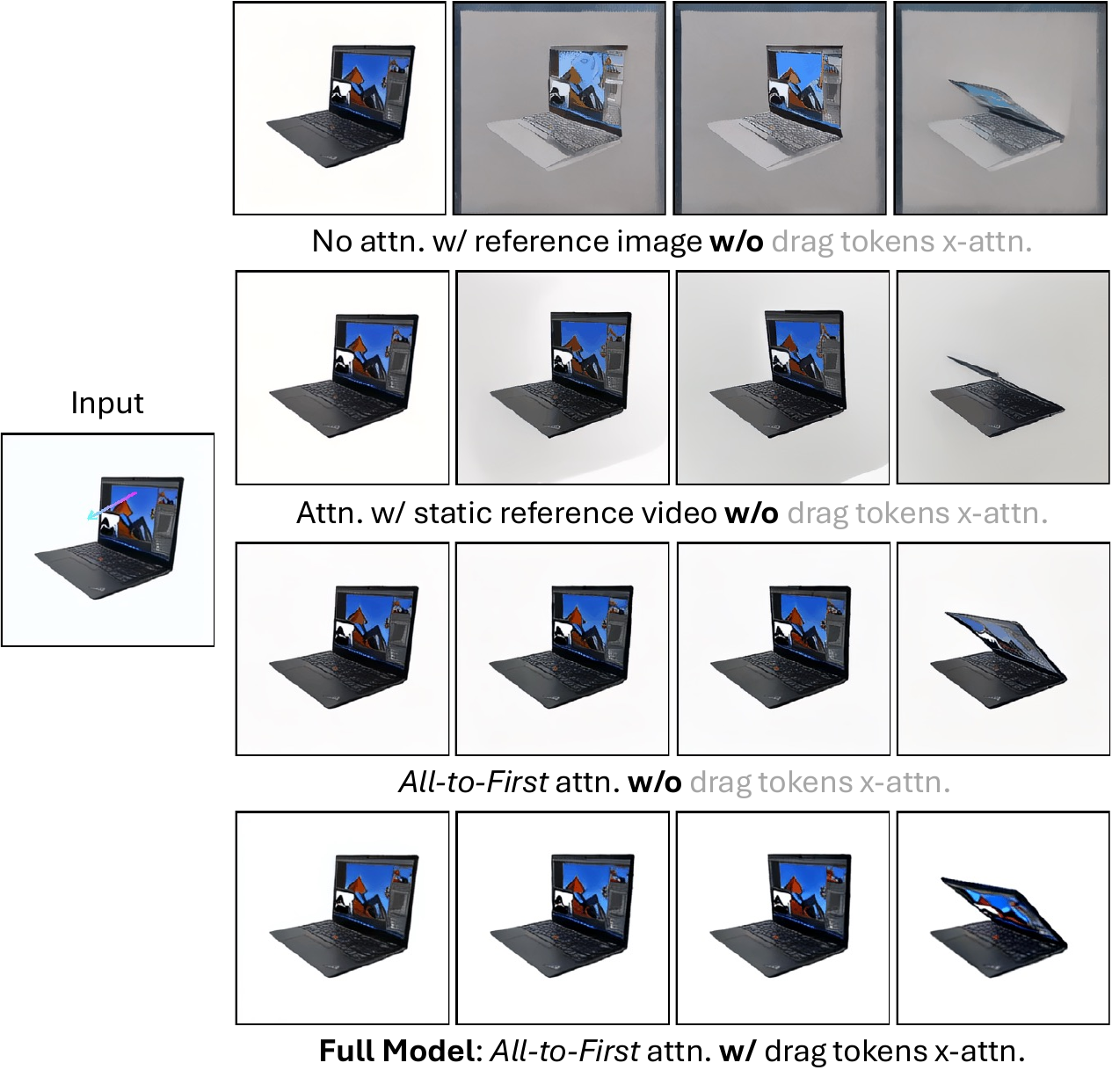}
    \caption{\textbf{Visualization} of samples generated by different model designs, where we show the last frame and the first three frames.
    While all designs produce nearly perfect first frames,
    our proposed \emph{all-to-first} attention module significantly enhances sample quality.
    Without this module, the generated samples often exhibit sub-optimal appearances and backgrounds.
    The cross-attention module with drag tokens further improves the appearance details.
    }%
    \label{fig:ablations}
\end{figure}
\begin{table}[tb!]
    \tablestyle{1.0pt}{1.0}
    \centering
    \renewcommand{\arraystretch}{1.0}
    \begin{tabular}{@{}l ccc ccc @{}}
        \toprule
        Setting & PSNR$\uparrow$ & SSIM$\uparrow$ & LPIPS$\downarrow$ & FVD$\downarrow$ & ME$\downarrow$ & \%WD$\downarrow$ \\
        \midrule
        \textbf{Drag conditioning} & & & & & & \\
        $\mathbb{A}$: Shift only w/o end loc. & 13.23 & 0.816 & 0.446 & 975.16 & 
        15.6 & $\geq 5$ \\
        $\mathbb{B}$: Shift+scale w/o end loc. & 22.98 & 0.917 & 0.093 & 223.20 & 
        \textbf{9.3} & 4 \\
        $\mathbb{C}$: Shift+scale w/ end loc. & 23.67 & 0.926 & 0.080 & 205.40 & 
        10.5 & 4 \\
        $\mathbb{D}$: $\mathbb{C}$ + x-attn.~w/ drag tok. & \textbf{24.00} & \textbf{0.929} & \textbf{0.069} & \textbf{170.43} & 
        9.8 & \textbf{1} \\
        \midrule
        \textbf{Attn.~w/ ref.~image} & & & & & & \\
        No attn. & 11.96 & 0.771 & 0.391 & 823.00 & 
        12.4 & $\geq 3$ \\
        Attn.~w/ static ref.~video & 17.51 & 0.874 & 0.233 & 483.18 & 
        13.6 & $\geq 8$ \\
        \emph{All-to-first} attn. & \textbf{23.67} & \textbf{0.926} & \textbf{0.080} & \textbf{205.40} & 
        \textbf{10.5} & 4 \\
        \bottomrule
    \end{tabular}
    \caption{\textbf{Ablations}.
    In addition to the standard metrics and motion error (\textbf{ME}) which we introduced in~\cref{sec:exp_setting},
    we also manually count the frequency of generated videos whose motion directions are opposite to the intention of their drag inputs (\% wrong direction, or \textbf{\%WD} in short).
    Here, $\geq$ indicates there are video samples whose motion directions are hard to distinguish.
    When ablating various designs of attention with the reference image, we use $\mathbb{C}$ as the base drag conditioning architecture.
    }%
    \label{tab:ablations}
\end{table}

\subsection{Ablations}%
\label{sec:exp_ablations}

We conduct ablations to analyze \method.
For each design choice, we train a separate model using the training split of the Drag-a-Move dataset with a batch size of $8$ for $30$k steps and evaluate on $100$ videos from its test split.
Results are shown in \cref{tab:ablations,fig:ablations} and discussed below.

\paragraph{Drag conditioning.}

\Cref{tab:ablations} compares \method with several variants of conditioning mechanisms (\cref{sec:method_drag}).
Adaptive normalization modules ($\mathbb{A}$ \vs $\mathbb{B}$) significantly improve both appearance quality (PSNR) and motion consistency (motion error ME).
Additionally, we perform an ablation study on the impact of drag encoding with the final termination location $v_k^N$ ($\mathbb{B}$ \vs $\mathbb{C}$). 
Providing the final motion destination of each drag as context for each frame proves beneficial.
Incorporating drag tokens in the cross-attention modules enhances spatial awareness and is effective ($\mathbb{C}$ \vs $\mathbb{D}$ and \cref{fig:ablations}).
Notably, by combining these (\ie, row $\mathbb{D}$), the model achieves a negligible rate of generated samples with incorrect motion directions.

\paragraph{Attention with the reference image.}

An evaluation of our proposed \emph{all-to-first} attention is shown in \cref{tab:ablations} and \cref{fig:ablations}.
We find that \emph{all-to-first attention} (\cref{sec:method_attn}) is essential for video quality.
We also compare \emph{all-to-first} attention with an alternative implementation inspired by the X-UNet design of~\cite{watson2023novel}, where we pass a static video consisting of the reference image copied $N$ times to the same network architecture and implement cross-attention between the clean (static) reference video branch and the noised video branch.
The latter strategy performs worse.
We hypothesize that this is due to distribution drift between the two branches, which forces the optimization to modify the pre-trained SVD's internal representations too much.

\section{Conclusion}%
\label{sec:conclusion}

We have introduced \method, a video generator that enables control of object motion at the part level via a set of sparse drags.
Compared to related works, \method incorporates several architectural innovations, including adaptive layer normalization modules, cross-attention modules with drag tokens, and all-to-first spatial attention modules.
Ablation studies demonstrate the effectiveness of these contributions.
\method is trained on \datasetF, a newly curated dataset of part-level object animations that we also contribute.
\method achieves state-of-the-art performance on several benchmarks and exhibits strong \emph{zero-shot} generalization to real-world cases.
It also demonstrates the viability of using video generators as proxies for learning a foundation model of the internal dynamics of objects.

\paragraph*{Acknowledgments.}

This work is in part supported by a Toshiba Research Studentship, EPSRC
SYN3D EP/Z001811/1, and ERC-CoG UNION 101001212.
We thank Luke Melas-Kyriazi, Jinghao Zhou, Minghao Chen and Junyu Xie for useful discussions.



{
\small
\bibliographystyle{ieee_fullname}
\bibliography{main}
}

\newpage

\appendix
\setcounter{section}{0} 
\renewcommand{\thesection}{\Alph{section}}

\section{Additional Details of the Drag Encoding}%
\label{sec:supp_enc}

Here, we give a formal definition of $\mathrm{enc}(\cdot, s)$ introduced in~\cref{sec:method_drag}. Recall that $\mathrm{enc}(\cdot, s)$ encodes each drag $d_k\coloneq (u_k, v_k^{1:N})$ into an embedding of shape $N\times s\times s\times 6$.
For each frame $n$, the first, middle, and last two channels (of the $c=6$ in total) encode the spatial location of $u_k$, $v_k^n$, and $v_k^N$, respectively.
Formally, $\mathrm{enc}(d_k, s)[n, \mathtt{:, :, :2}]$ is a tensor of all negative ones except for
$
\mathrm{enc}(d_k, s)[n, \left\lfloor\frac{s\cdot h}{H}\right\rfloor, \left\lfloor\frac{s\cdot w}{W}\right\rfloor,\mathtt{:2}]
=
\left(
  \frac{s\cdot h}{H} -
  \left\lfloor \frac{s\cdot h}{H} \right\rfloor,
  \frac{s\cdot w}{W} -
  \left\lfloor \frac{s\cdot w}{W} \right\rfloor
\right)
$
where $u_k = (h, w) \in \Omega = \left \{1, \ldots, H\right \}\times \left \{1, \ldots, W\right \}$.
The other $4$ channels are defined similarly, with $u_k$ replaced by $v_k^n$ and $v_k^N$.

\section{Additional Details of Data Curation}

\subsection{Implementation Details}
We use the categorization provided by GObjaverse~\cite{qiu2024richdreamer} and exclude 3D models classified as `\texttt{Poor-Quality}' as a pre-filtering step prior to our proposed filtering pipelines (\cref{sec:dataset}).

When using GPT-4V to filter \dataset into \datasetF, we designed the following prompt to cover a wide range of cases to be excluded:

\tcbset{colback=lightgray, colframe=lightgray, boxrule=0pt, sharp corners}
\begin{tcolorbox}
\textbf{System}: You are a 3D artist, and now you are being shown some animation videos depicting an animated 3D asset. You are asked to filter out some animations. 
    
You should filter out the animations that:

(1) have trivial or no motion, i.e., the object is simply scaling, rotating, or moving as a whole without part-level dynamics;

or (2) depict a scene and only a small component in the scene is moving;

or (3) have motion that is imaginary, i.e., the motion is not the usual way of how the object moves and it's hard for humans to anticipate;

or (4) have very large global motion so that the object exits the frame partially or fully in one of the frames;

or (5) have changes in object color that are not due to lighting changes;

or (6) have motion that causes different parts of the same object to disconnect, overlap in an unnatural way, or disappear;

or (7) have motion that is very chaotic, for example objects exploding or bursting apart.

\textbf{User}: For the following animation (as frames of a video), \texttt{frame1}, \texttt{frame2}, \texttt{frame3}, \texttt{frame4}, tell me, in a single word `Yes' or `No', whether the video should be filtered out or not.
\end{tcolorbox}

The cost of GPT-4V data filtering is about \$500.

\begin{figure}[H]
\centering
\includegraphics[width=\linewidth]{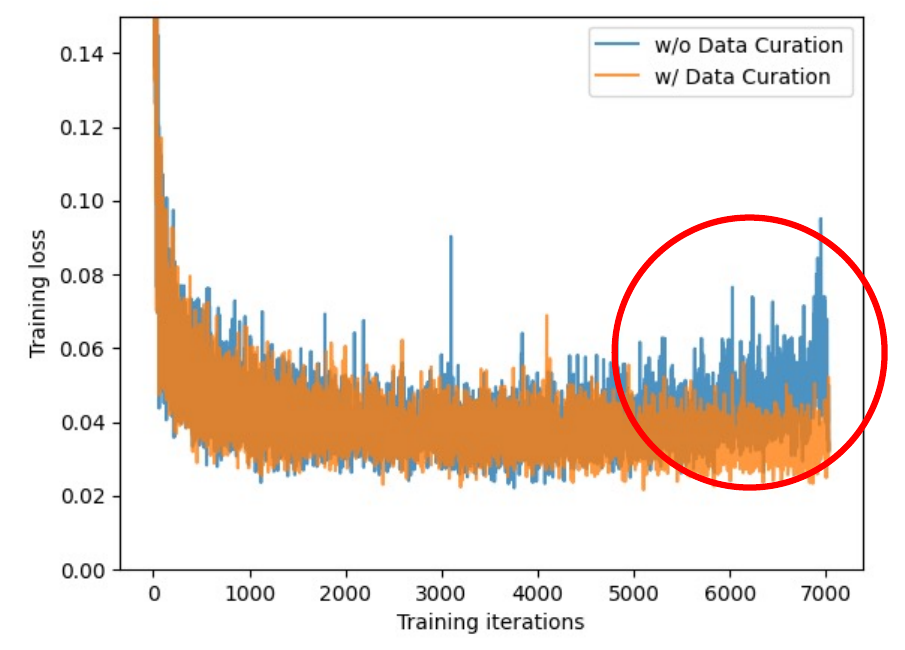}
\caption{Data curation helps stabilize training.}%
\label{fig:curation}
\end{figure}

\begin{table}[H]
\tablestyle{3.0pt}{1.0}
\centering
\begin{tabular}{c cccc}
    \toprule
    Setting & PSNR$\uparrow$ & SSIM$\uparrow$ & LPIPS$\downarrow$ & FVD$\downarrow$ \\
    \midrule
    w/o Data Curation & 6.04 & 0.411 & 0.703 & 1475.35 \\
    w/ Data Curation & \textbf{19.87} & \textbf{0.884} & \textbf{0.181} & \textbf{624.47} \\
    \bottomrule
\end{tabular}
\caption{Training on more abundant but lower-quality data leads to lower generation quality. Here, `w/o Data Curation' model is trained on \dataset while `w/ Data Curation' model is trained on \datasetF. Both models are trained for $7$k iterations. Evaluation is performed on the test split of Drag-a-Move.}%
\label{tab:curation}
\end{table}

\subsection{Less is More: Data Curation Helps at Scale}%
\label{sec:exp_data}

To verify that our data curation strategy from~\cref{sec:dataset} is effective, we compare two models trained on \dataset and \datasetF, respectively, under the same hyperparameter setting.
The training dynamics are visualized in~\cref{fig:curation}.
The optimization collapses towards $7$k iterations when the model is trained on a less curated dataset, resulting in much lower-quality video samples (\cref{tab:curation}).
This suggests that when fine-tuning a pre-trained video diffusion model to generate part-level motion, the quality of the data is more critical than its quantity.

\section{Additional Experiment Details}%
\label{sec:supp_exp}

\subsection{Training Details}

\paragraph{Data.}
Our final model is fine-tuned on the combined dataset of Drag-a-Move~\citep{li2024dragapart} and \datasetF (\cref{sec:dataset}).
During training, we balance various types of part-level dynamics to control the data distribution.
We achieve this by leveraging the categorization provided by GObjaverse~\citep{qiu2024richdreamer} and sampling individual data points with the following hand-crafted distribution: $p($Drag-a-Move$)=0.3$, $p($\datasetF, 
category `\texttt{Human-Shape}'$)=0.25$, $p($\datasetF, 
category `\texttt{Animals}'$)=0.25$, $p($\datasetF, 
category `\texttt{Daily-Used}'$)=0.05$, $p($\datasetF, 
other categories$)=0.15$.

\paragraph{Architecture.} We zero-initialize the final convolutional layer of each adaptive normalization module before fine-tuning. With our introduced modules, the parameter count increases to $1.68$B from the original $1.5$B in SVD\@.

\paragraph{Training.} We fine-tune the base SVD on videos of $256\times 256$ resolution and $N=14$ frames with a batch size of $64$ for $12,500$ iterations. We adopt SVD's continuous-time noise scheduler, shifting the noise distribution towards more noise with
$
\log \sigma \sim \mathcal{N}(0.7, 1.6^2)
$,
where $\sigma$ is the continuous noise level following the presentation in~\citep{blattmann2023stable}.
Training takes roughly $10$ days on a single Nvidia A6000 GPU, where we accumulate gradients for $64$ steps.
We enable classifier-free guidance (CFG)~\citep{ho2022cfg} by randomly dropping the conditional drags $\mathcal{D}$ with a probability of $0.1$ during training.
Additionally, we track an exponential moving average of the weights at a decay rate of $0.9999$.

\subsection{Inference and Evaluation Details}

\paragraph{Inference.}
Unless stated otherwise, samples are generated using $S=50$ diffusion steps.
We adopt linearly increasing CFG~\citep{blattmann2023stable} with a maximum guidance weight of $5.0$.
Generating a single video takes roughly $20$ seconds on an Nvidia A6000 GPU\@.

\paragraph{Baselines.}
For DragNUWA~\citep{yin2023dragnuwa}, DragAnything~\citep{wu2024draganything}, and Image Conductor~\citep{li2024imageconductor}, we use their publicly available checkpoints.
DragNUWA and DragAnything operate at a resolution of $576\times 320$, and Image Conductor at $384\times 256$.
Following previous work~\citep{li2024dragapart}, we first pad the square input image $y$ along the horizontal axis to the correct aspect ratio and resize it to the corresponding resolution, then remove the padding from the generated frames and resize them back to $256\times 256$.
For methods that require text prompts (\ie, DragNUWA and Image Conductor), we use generic prompts to describe the category of the evaluation images (\eg, `\texttt{A Furniture}' for Drag-a-Move and `\texttt{A person}' for Human3.6M).
Note that Image Conductor is trained on $16$-frame videos instead of $14$-frame ones.
We experimented with (1) simply generating $14$ frames at inference time; and (2) generating $16$ frames and discarding the last two frames. The latter gives slightly better results, which we report.
We find that tasking it to generate $14$-frame videos produces reasonable results which we report.
All metrics are computed on $14$-frame videos of resolution $256\times 256$.

We train DragAPart~\citep{li2024dragapart} for $100$k iterations using its official implementation on the same combined dataset of Drag-a-Move and \datasetF used for training \method.
Since DragAPart is an image-to-image model, we independently generate $14$ frames conditioned on gradually extending drags to obtain the video.

For Sora~\cite{brooks24video}, we uploaded the conditioning image in~\cref{fig:comparison} as the start frame. Since the model does \emph{not} support motion control, we
manually crafted the following prompt to convey the motion condition:

\begin{tcolorbox}
A photorealistic video of a modern, light grey wooden sideboard with a natural wood top.
The three drawers at the top remain completely static and closed throughout the entire video, without any movement or displacement. From this initial state, only the bottom cabinet doors begin to slowly and smoothly close, moving in a natural, physically plausible manner. The motion follows proper hinge mechanics, ensuring perfect alignment, symmetry, and realism, with no jerky or unnatural movement. The camera remains fixed in the same frontal view, maintaining the exact perspective of the reference image. The lighting is soft and even, enhancing the wood texture, clean lines, and elegant design without casting harsh shadows or introducing distractions. The video maintains a high-quality, cinematic appearance, with no additional objects or background elements.
\end{tcolorbox}

\section{Video Diffusion Models on Out-of-Domain Resolutions}%
\label{sec:supp_svd}

\begin{figure}[htb!]
    \centering
    \includegraphics[width=\columnwidth]{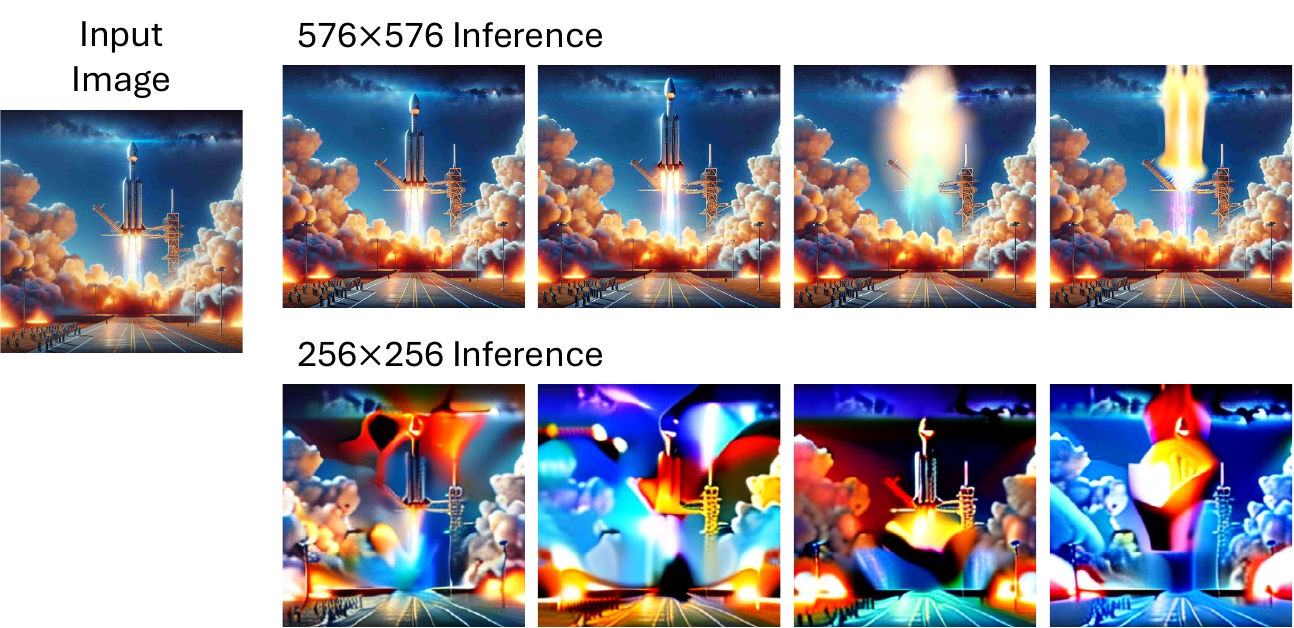}
    \caption{Stable Video Diffusion \emph{fails} to generalize robustly to out-of-domain resolutions at inference time.}%
    \label{fig:svd-illustration}
\end{figure}

The convolution and attention modules in video diffusion models like SVD are \emph{not} invariant to input resolution.
As demonstrated in~\cref{fig:svd-illustration}, our base model SVD, which was trained on videos with resolution $1024\times 576$, \emph{cannot} generate high-quality videos at out-of-domain resolutions such as $256\times 256$.
We hypothesize that this resolution shift makes fine-tuning susceptible to local optima, resulting in visually cluttered generations (\cref{fig:ablations}). All-to-first attention (\cref{sec:method_attn}) significantly reduces this appearance degradation.

\section{Discussions}

\begin{figure}[tb!]
    \centering
    \includegraphics[width=\columnwidth]{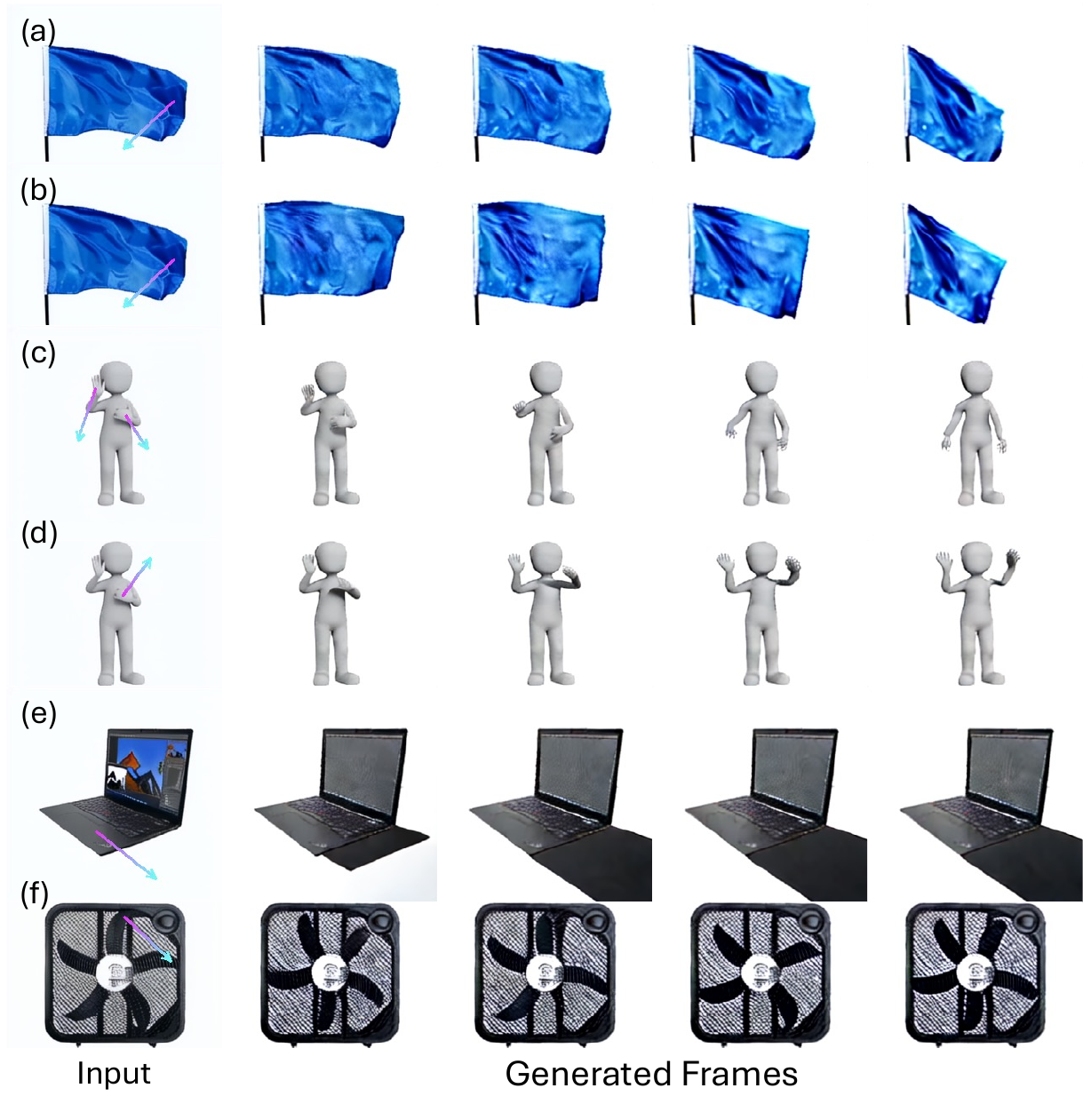}
    \caption{\textbf{More examples} generated by \method.}%
    \label{fig:more-examples}
\end{figure}

\begin{figure}[t!]
\small
\setlength{\abovecaptionskip}{2pt} 
\centering
\includegraphics[width=\columnwidth]{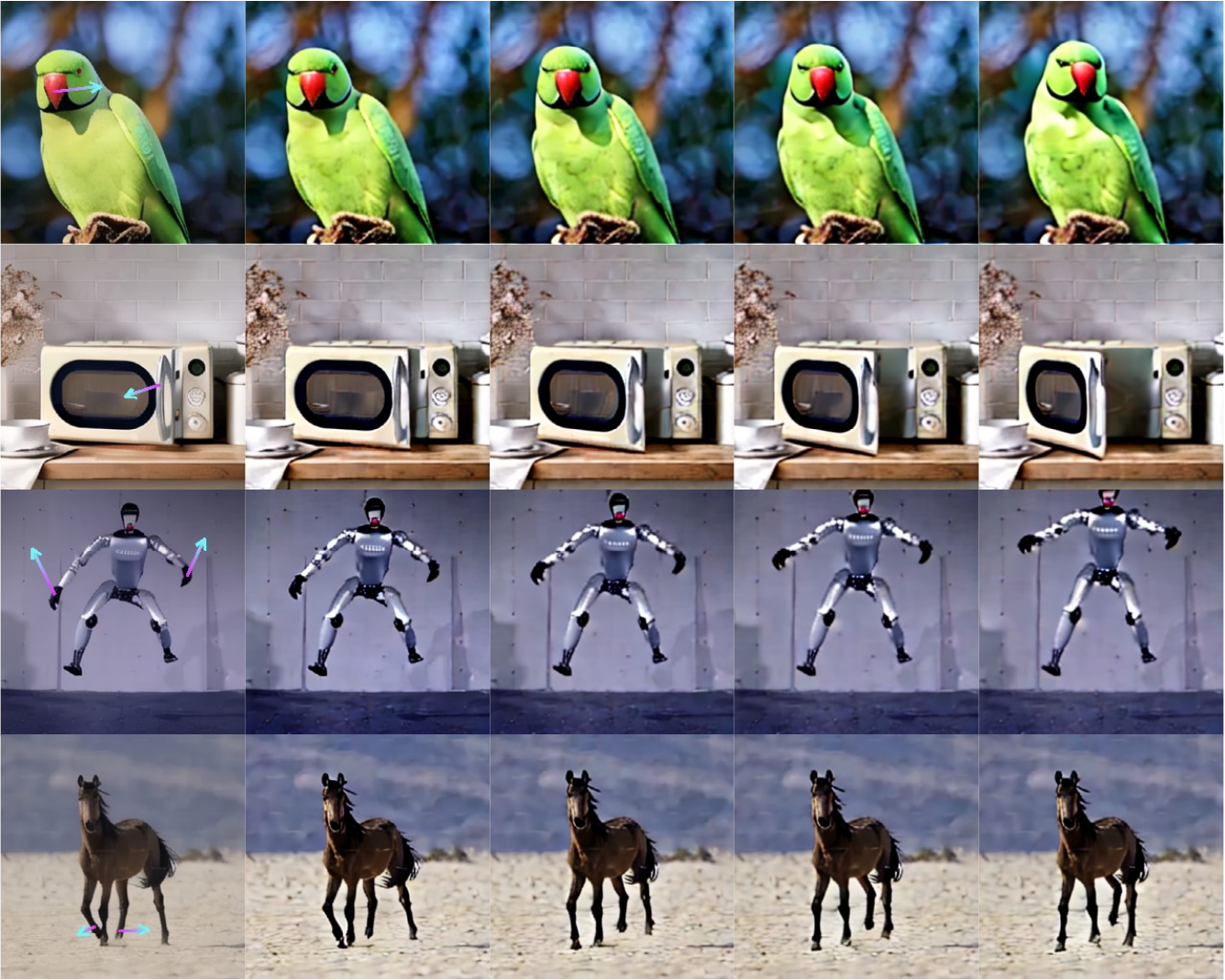}
\caption{\textbf{Results} on images with diverse backgrounds.}%
\label{fig:supp-results-real}
\end{figure}

\paragraph{Motion diversity.}

In~\cref{fig:more-examples}(a-d), we show that \method can generate diverse part-level animations, both across different random seeds when conditioned on the same input image and set of drags (\ie, a and b), and across different sets of drags when conditioned on the same input image (\ie, c and d).

\paragraph{Part-level \vs object-level motion.}

In this work, we focus on synthesizing \emph{internal}, \emph{part-level} motion. 
To achieve this, we curated \datasetF to specifically learn motions involving object parts being manipulated.
As a result, \method is not designed for \emph{global} object motion and may produce artifacts when the input drag(s) do \emph{not} correspond to meaningful part-level movement (\cref{fig:more-examples}e).

\paragraph{Failure cases.}

\method may fail to maintain the shape of objects, occasionally leading to the disappearance of certain parts.
This issue is particularly evident when physically plausible motion necessitates precise coordination among multiple object parts, such as the five fan blades in~\cref{fig:more-examples}f.

\paragraph{Results with real-world backgrounds.}
Although all training frames are rendered with a white background, \method retains some ability from the SVD backbone to handle complex backgrounds, as illustrated in~\cref{fig:supp-results-real}.
Better results could be obtained by incorporating, \eg, random backgrounds during training.

\paragraph{Limitations.}

Another limitation of our model is its slight difficulty in preserving the exact color appearance of objects during inference on real-world images.
This issue arises due to two primary factors:
(1) the synthetic 3D models in \datasetF typically feature high-contrast, stylized textures, leading to a train-test discrepancy in color distributions;
and (2) when testing at a lower resolution (\eg, $256\times 256$) compared to the native resolution of SVD, noise in the denoiser's output can propagate across a larger region of the image because of the fixed receptive field of convolutional layers, leading to many instances having a slightly flickering appearance.

\paragraph{Future work.}

While most motion-conditioned video generators prioritize object-level motion over fine-grained part-level motion, we have demonstrated it is feasible to learn a part-level motion prior using a modestly sized, high-quality synthetic dataset that generalizes effectively to real-world data.
Future research may develop a dynamic routing mechanism that integrates both part-level and object-level dynamics.

\end{document}